\documentclass[letterpaper]{article} % DO NOT CHANGE THIS
\usepackage{aaai25}  % DO NOT CHANGE THIS
\usepackage{times}  % DO NOT CHANGE THIS
\usepackage{helvet}  % DO NOT CHANGE THIS
\usepackage{courier}  % DO NOT CHANGE THIS
\usepackage[hyphens]{url}  % DO NOT CHANGE THIS
\usepackage{graphicx} % DO NOT CHANGE THIS
\usepackage{natbib}  % DO NOT CHANGE THIS AND DO NOT ADD ANY OPTIONS TO IT
\usepackage{caption} % DO NOT CHANGE THIS AND DO NOT ADD ANY OPTIONS TO IT
\usepackage{algorithm}
\usepackage{algorithmic}

\newtheorem{definition}{Definition}

\usepackage{booktabs}
\usepackage{amsmath}
\usepackage{bm}
\usepackage{subfigure}
\usepackage{graphicx}
\usepackage{subcaption}
\usepackage{xcolor}

\usepackage{newfloat}
\usepackage{listings}
\DeclareCaptionStyle{ruled}{labelfont=normalfont,labelsep=colon,strut=off} % DO NOT CHANGE THIS
\floatstyle{ruled}
\newfloat{listing}{tb}{lst}{}
\floatname{listing}{Listing}
\title{Mixture of Experts Based Multi-Task Supervise Learning from Crowds}
\author{
    Tao Han,
    Huaixuan Shi,
    Xinyi Ding,
    Xi-Ao Ma,
    Huamao Gu,
    Yili Fang\footnote{Corresponding author}
}
\affiliations{
    School of Computer Science and Technology, Zhejiang Gongshang University, Hangzhou 310018, China \\
    \{hantao, xding, mxa, ghmsjq, fangyl\}@zjgsu.edu.cn, 22020100076@pop.zjgsu.edu.cn
}

\usepackage{bibentry}
\begin{document}

\maketitle

\begin{abstract}
%Existing truth inference methods in crowdsourcing aim to map redundant labels and items to the ground truth. 
Existing learning-from-crowds methods aim to design proper aggregation strategies to infer the unknown true labels from noisy labels provided by crowdsourcing.
They treat the ground truth as hidden variables and use statistical or deep learning based worker behavior models to infer the ground truth. However, worker behavior models that rely on ground truth hidden variables overlook workers' behavior at the item feature level, leading to imprecise characterizations and negatively impacting the quality of learning-from-crowds.
This paper proposes a new paradigm of multi-task supervised learning-from-crowds, which eliminates the need for modeling of items's ground truth in worker behavior models. Within this paradigm, we propose a worker behavior model at the item feature level called Mixture of Experts based Multi-task Supervised Learning-from-Crowds (MMLC), then, two aggregation strategies
are proposed within MMLC. The first strategy, named MMLC-owf, utilizes clustering methods in the worker spectral space to identify the projection vector of the oracle worker. Subsequently, the labels generated based on this vector are regarded as the items's ground truth
The second strategy, called MMLC-df, employs the MMLC model to fill the crowdsourced data, which can enhance the effectiveness of existing  aggregation strategies
. Experimental results demonstrate that MMLC-owf outperforms state-of-the-art methods and MMLC-df enhances the quality of existing learning-from-crowds methods.
%truth inference 
\end{abstract}

\section{Introduction}
%Truth inference 
Existing learning-from-crowdsmethods can be broadly classified into two categories: weakly supervised and supervised approaches. In the weakly supervised approach, unknown ground truth are treated as hidden variables. This involves utilizing statistics from workers' noisy answers to calculate results directly. Alternatively, it entails creating worker behavior models and employing unsupervised learning methods such as the EM algorithm to estimate unknown parameters and infer the ground truth. The weakly supervised approach can further be classified into statistical learning and deep learning methods based on whether considering the features of items. Statistical learning methods, such as MV~\cite{imamura2018analysis}, DS~\cite{dawid1979maximum}, and HDS~\cite{karger2011iterative,li2014error}, do not incorporate item features. In contrast, deep learning methods like Training Deep Neural Nets~\cite{gaunt2016training} take item features into account. In the supervised approach, a classifier model is first constructed with item features as the input and ground truth as the output. Then, a worker behavior model is created based on a confusion matrix that establishes the relationship between one item's ground truth and the worker labels. On this basis, supervised learning is implemented using the classifier model and the worker behavior model by treating the worker labels as supervisory information. Finally, the output of the classifier model is used as the inferred ground truth. In recent years, various learning-from-crowds
%truth inference 
methods based on supervised learning have been proposed, such as Crowdlayer\cite{rodrigues2018deep}, CoNAL\cite{chu2021learning}, and UnionNet\cite{wei2022deep}. However, the worker behavior model based on the confusion matrix faces challenges in effectively capturing variations in feature characteristics across different items. Neglecting these variations in worker behavior under different conditions can result in inaccurate representations of worker behavior, consequently impacting the quality of truth inference. For example, in handwritten digit recognition, workers generally have high accuracy. Suppose there are two items: one closely resembles the digit ``1,'' but its ground truth is actually ``7,'' and the other is a normal ``7.'' The former receives many labels as ``1,'' while the latter rarely gets labeled as ``1.'' Under the worker behavior model based on the confusion matrix, it is difficult to model the labeling behavior of such difficult items accurately. Therefore, there is a high probability that the model will interpret the former with ``1'' as the ground truth, leading to incorrect judgments. The quality of  aggregation strategies
%truth inference 
is influenced by uncertainty from hidden variables, the method's data adaptability, and the accuracy in characterizing worker behavior. The purpose of this paper is to develop a supervised model that can achieve high-quality truth inference with a worker behavior model at the item feature level.

In this paper, we propose Multi-task Supervised Learning-from-crowds (MLC), a novel paradigm for crowdsourcing learning. Unlike the traditional paradigm, MLC does not rely on the ground truth of the items but instead focuses on understanding the unique behavior of individual workers across different items. When multiple workers handle the same item, they share the item's features, leading to a multi-task learning paradigm. Within this paradigm, we propose a method called Mixture of Experts-based Multi-task Supervised learning-from-crowds (MMLC). MMLC does not utilize a single classifier but instead creates multiple expert modules. The outputs from these expert modules serve as the \emph{bases} of the worker spectral space. Each worker is represented by his or her projection vector in the spectral space that characterizes their behavior. The worker behavior model provides a more precise depiction of their behavior across different items by accurately modeling the workers' behavior on item features. However, it is important to note that the model itself cannot determine the ground truth. To address this limitation, we introduce two aggregation strategies based on MMLC. The first strategy involves analyzing the distribution of workers' projections in the worker spectral space. We identify the projection of the oracle worker by applying clustering methods, and consider its labels as the ground truth. This approach is referred to as Oracle Worker Finding of MMLC (MMLC-owf). The second strategy leverages the sparsity of crowdsourced data to fill the original dataset with MMLC outputs, generating a new crowdsourcing dataset. Existing learning-from-crowds methods can then be applied within this framework, which is called Data Filling of MMLC (MMLC-df). The main contributions are as follows:
\begin{itemize}
    \item We introduce a novel paradigm of multi-task supervised learning-from-crowds and propose a novel worker behavior model called MMLC based on feature-level behavior modeling.
    \item We leverage MMLC to identify the oracle worker for labeling items as the ground truth, referred to as MMLC-owf. Experimental results demonstrate that the labels obtained using this method exhibit higher quality compared to state-of-the-art methods.
    \item We introduce an aggregation framework called MMLC-df, which leverages the MMLC model to fill sparse crowdsourced data. This framework then applies aggregation
    %truth inference 
    methods to determine the ground truth. Experimental results demonstrate that MMLC-df significantly enhances learning-from-crowds
    %truth inference 
    methods, leading to higher quality results.
\end{itemize}

\section{Related Work}

\textbf{Weakly supervised approaches}: These approaches model the relationship between workers' noisy responses and the true labels, treating the ground truth as a latent variable and employing weak supervision techniques to deduce it. MV~\cite{sheng2017majority} is a commonly used method that assumes the most frequent response as the ground truth, but it fails to account for worker variability. To address this limitation, \cite{tao2020label} proposed a model that separately considers the majority and minority responses, factoring in the labeling quality of workers. The DS~\cite{dawid1979maximum} uses a confusion matrix to characterize worker behavior and estimates parameters with the EM algorithm to infer the ground truth. HDS~\cite{raykar2010learning} posits equal chances of erroneous worker choices, refining the confusion matrix with this assumption. GLAD~\cite{whitehill2009whose}, conversely, factors in both the proficiency of workers and the inherent challenge of tasks, utilizing a sigmoid function to depict worker behavior and the EM algorithm to ascertain the ground truth.  Various weakly supervised learning-from-crowds methods integrate deep learning to deduce the ground truth, initiating with strategies that aggregate the noisy labels into an initial answer table. Post this, a neural network is trained for classification, leveraging the curated label set. 
While these methods~\cite{gaunt2016training,ghiassi2022labnet,zhu2023improving} propose label reliability metrics that significantly influence the outcomes of crowd-based learning. Additionally, \cite{xu2024crowdsourcing} presents a two-stage method that uses a multi-centroid grouping penalty to incorporate subgroup structures for tasks and workers in inferring the ground truth. While weakly supervised methods have seen successes, they are often hampered by sparse data and the treatment of ground truth as a hidden variable, which limits their accuracy.

\noindent\textbf{Supervised approaches:} These methods focus on creating a classifier to link item features with the ground truth and a worker behavior model using a confusion matrix to reflect the relationship between true labels and worker responses. These responses act as supervision, allowing for joint training of both models in a supervised manner, with the classifier's output serving as inferred ground truth\cite{chu2021improve,ibrahim2023deep}. Techniques like the Expectation-Maximization (EM) algorithm are used for label aggregation and classifier training. Crowdlayer\cite{rodrigues2018deep} replaces the traditional confusion matrix with a crowd layer, integrating label reasoning and classifier training for more accurate results. Tan and Chen\cite{tanno2019learning} enhance model accuracy by applying confusion and labeled transfer matrices alongside classifier predictions. Other approaches\cite{gao2022learning,cao2023learning} introduce worker weight vectors and focus on modeling label reliability to estimate worker abilities. UnionNet\cite{wei2022deep} aggregates worker annotations into a parameter transfer matrix to facilitate training. The CoNAL method\cite{chu2021learning} categorizes noise into common and individual types, effectively managing diverse labeling noise. 
Despite its strengths, supervised learning faces challenges in precisely characterizing worker behavior, making it difficult to achieve optimal results when candidate answers are poorly differentiated.

Departing from the traditional latent variable approach to ground truth, our method concentrates on the interplay between worker behavior and item features. This shift enables us to develop a supervised learning framework from the crowd, yielding a nuanced worker behavior model. With this model, we pinpoint the oracle workers capable of precise labeling, thus ascertaining the ground truth. 

%Additionally, the model is leveraged to enrich sparse crowdsourced data, after which conventional aggregation techniques are applied to derive the ground truth from the enriched dataset.

\section{Problem Formulation}

 Our main goal is to obtain a worker behavior model and achieve joint learning of worker abilities by utilizing multi-task learning from crowds to aggregate the ground truth. Let $\mathcal{W}=\{w_j\}$ denote the worker set, where $w_j$ represents an individual worker, and $\mathcal{X}=\{x_i\}$ denote the set of items, where $x_i$ represents a single item to be labeled. The labels for each item belong to the category set $\mathcal{K} = \{k\}$. We use $y_{ij}$ to denote the category label assigned by worker $w_j$ to item $x_i$. We have an indicator function $I_{ij}$, where $I_{ij}=1$ if $y_{ij}$ exists and $I_{ij}=0$ otherwise. Consequently, we obtain the crowdsourced triples dataset $\mathcal{D}=\{\langle x_i, w_j, y_{ij}\rangle |I_{ij}=1\}$. With regards to learning from crowd in crowdsourcing, we provide the following definition:

\begin{definition}\textbf{\emph{(Problem of Learning-from-Crowds (LC Problem)}}
By modeling and learning from the crowdsourced label dataset $\mathcal{D}$, the problem of learning-from-crowds aims to find a function $g^*: \mathcal{X} \rightarrow \mathcal{K}$ such that
\begin{equation}
    g^* = \arg\min_{g\in \mathcal{H}}\sum_{i=1}^{|\mathcal{X}|} \mathcal{L}\left( \hat{z}_i, g(x_i) \right) + \lambda \|g\|_\mathcal{H}.
\end{equation}
\end{definition}
Here, $\mathcal{H}$ denotes the hypothesis space of functions, $\|\cdot\|_\mathcal{H}$ denotes the norm of hypothesis space, $\lambda$ is the regularization coefficient, $\mathcal{L}$ is the loss function, and $\hat{z}_i= t_i(\mathcal{D})$ is the estimation of label $z_i$ for item $x_i$ from learning on the dataset. Since the crowdsourced LC problem is an unsupervised learning-from-crowds problem without supervised information, the estimation of ground truth is utilized instead of the goal of learning.

\begin{definition}\textbf{\emph{(Problem of Multi-task supervise Learning-from-Crowds (MLC Problem))}}
Let $\mathcal{S}_j=\{(x_i , y_{ij})\}_{x_i\in \mathcal{X}_j}$ denote the crowdsourced training dataset for worker $w_j$, where $\mathcal{X}_j=\{x_i|I_{ij}=1\}$.
The labels provided by worker $j$ can be regarded as the $j$-th task for the corresponding item. Consequently, we obtain the dataset  as $\mathcal{S}=\bigcup_j \mathcal{S}_j$. The problem of multi-task supervised learning-from-crowds is to find a worker behavior function $f^*\in \mathcal{H}$ such that
\begin{equation}
    f^* = \arg\min_{f\in \mathcal{H}} \sum_{w_j\in\mathcal{W}} \frac{1}{|\mathcal{X}_j|} \sum_{x_i\in \mathcal{X}_j} \mathcal{L}\left(y_{ij}, f_{w_j}(x_i)\right) + \lambda \|f\|_\mathcal{H},
\end{equation}
\end{definition}
where $\mathcal{H}$ is a vector-valued reproducing kernel Hilbert space with functions $f:\mathcal{X}\rightarrow \mathcal{K}^{|\mathcal{W}|}$ having components $f_j:\mathcal{X}\rightarrow \mathcal{K}$ .

We can observe that the solution to the MLC problem does not directly address the LC problem. Therefore, we provide two approaches to tackle this issue. The first approach is to identify an oracle worker $w_{oracle}$ based on the distribution of workers. We then consider the labels provided by this oracle worker as the ground truth, that is,
\begin{equation}
    g^*(x_i) = f_{w_{oracle}}^*(x_i).
\end{equation}
The second approach considers the sparsity characteristic of crowdsourced data, where workers do not annotate every item. Consequently, we can utilize the results of MLC to generate a new dataset for inference. The new crowdsourced data can be defined as follows:
\begin{equation}
    \mathcal{D}' = \mathcal{D}\cup \left\{\langle x_i,w_j,\hat{y}_{ij}\rangle  \Big| \hat{y}_{ij} = f_{w_j}^*(x_i),I_{ij}=0\right\}.
\end{equation}

\section{Proposed Methodology}

% \begin{figure*}[t]
%     \centering
%     \includegraphics[width=0.7\textwidth]{figures/model2.png}
%     \caption{Model Structure of MMLC.}
%     \label{fig:model}
% \end{figure*}
\begin{figure}[t]
    \centering
    \includegraphics[width=1\columnwidth]{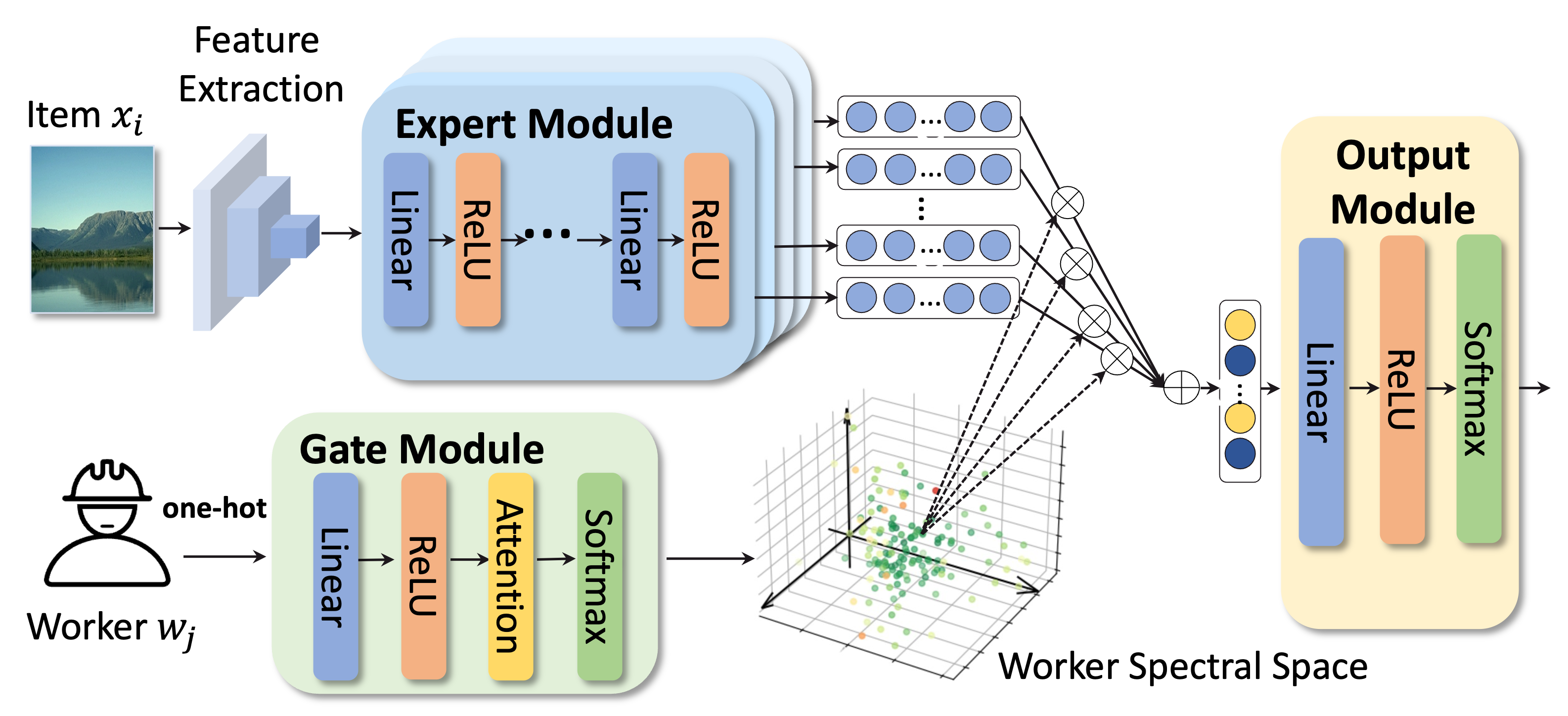}
    \caption{Model Structure of MMLC.}
    \label{fig:model}
\end{figure}
%%%%%%%%%%%%%%%%%%%%%%%%%%%%%%%%%%%%%%%%%
% 针对MLC Problem，本文提出了一种基于混合专家的模型(称为Mixture of Experts based Multi-task Supervise Learning from Crowds（MMLC）)，该模型利用混合专家的形式去刻画工人对于item特征不同的关注度，旨在捕获工人在面对不同item时特征级的表现差异。
% 模型的框架如图\ref{fig:model}所示，它由Expert Module, Gate Module， Output Module三个主要模块组成。
%%%%%%%%%%%%%%%%%%%%%%%%%%%%%%%%%%%%%%%%%
To address the MLC problem, we propose a Mixture of Experts based Multi-task Supervised Learning-from-Crowds (MMLC) model. This model utilizes mixture of experts to characterize the varying attention of workers towards different item features, aiming to capture the feature-level behavior differences of workers when dealing with various items. The framework of the model is shown in Fig.~\ref{fig:model}. It consists of three main modules: expert module, gate module, and output module.

%%%%%%%%%%%%%%%%%%%%%%%%%%%%%%%%%%%%%%%%%
% In Expert Module，item通过特征提取器得到item特征向量$x_{i}$，然后item特征向量进入m个专家模块，每个专家模块描述的是工人关注不同的特征信息所展现的不同工人行为特点。
% 每个专家模块对特征向量进行变换和压缩，最终得到专家模块的输出矩阵:
% \begin{equation}
%     \bm{U}(x_i) = \left(\bm{u}_1(x_i),\bm{u}_2(x_i),..., \bm{u}_E(x_i) \right).
% \end{equation}
% 每个专家子模块都是相同结构的多层全连接神经网络，每层均由ReLU作为激活函数.
% 对于每个专家模块$\bm{u}_e$，将高维特征向量$x_i$转换为低维向量$\bm{u}_e(x_i)$。
%%%%%%%%%%%%%%%%%%%%%%%%%%%%%%%%%%%%%%%%%
In the expert module, each item is processed by a feature extractor to obtain an item feature vector $x_{i}$. Then, the item feature vector is fed into $m$ expert modules, where each module captures the unique characteristics of worker behavior associated with different feature information. Each expert module performs transformations and compressions on the feature vector, resulting in the output matrix of the expert module: $\bm{U}(x_i) = \left(\bm{u}_1(x_i),\bm{u}_2(x_i),..., \bm{u}_E(x_i) \right)$.
Each expert sub-module follows the same structure, consisting of multiple layers of fully connected neural networks with ReLU activation functions in each layer.
For each expert sub-module $\bm{u}_e$, the high-dimensional feature vector $x_i$ is transformed into a low-dimensional vector $\bm{u}_e(x_i)$.

%%%%%%%%%%%%%%%%%%%%%%%%%%%%%%%%%%%%%%%%%
% In Gate Module, 该模块构建了一个门控网络去实现对专家模块的选择。门控网络输入是worker数据，输出的是工人在工人spectral space长度为$E$的projection vector。这个工人人特征空间的basis是专家子模块的输出。
% 具体而言，该模块将每个众包工人$w_j$ 经过One-Hot处理的向量作为输入,经过一层全连接ReLU层，一层Attention层，一层Softmax层，最后输出一个长度为$E$的工人projection vector$\bm{v}(w_j)=(v_1(w_j),v_2(w_j),...,v_E(w_j))^T$。工人$w_j$在专家子模块作为basis的工人spectral space投影表示为：
% \begin{equation}
%     proj_{\bm{U}(x_i)} (w_j) = \sum_{e=1}^E v_e (w_j) \bm{u}_e (x_{i}).
% \end{equation}
%%%%%%%%%%%%%%%%%%%%%%%%%%%%%%%%%%%%%%%%%
In the gate module, a gate network is constructed to control the selection of expert modules. This gate network takes worker data as input and generates a projection vector of the worker in the worker spectral space, with a length of $E$. The \emph{bases} of the worker spectral space are the outputs of the expert sub-modules.  Specifically, the module takes the one-hot encoded vector representing each worker $w_j$ as input. After passing through a fully connected ReLU layer, the data proceeds through an attention layer and a softmax layer. Finally, it produces a worker projection vector $\bm{v}(w_j)=(v_1(w_j),v_2(w_j),...,v_E(w_j))^T$ with a length of $E$. The projection of worker $w_j$ in the worker spectral space with the expert sub-modules as the \emph{bases} is:
\begin{equation}
    proj_{\bm{U}(x_i)} (w_j) = \sum_{e=1}^E v_e (w_j) \bm{u}_e (x_{i}).
\end{equation}
Here, the attention layer helps reduce the model's dependence on unimportant or redundant features, improving the model's efficiency and accuracy by focusing on the most useful information.

%%%%%%%%%%%%%%%%%%%%%%%%%%%%%%%%%%%%%%%%%
% In Output Module, 输出模块输出工人对item的标注结果。输出模块根据每个工人对专家模块的选择输出每个工人对该item的标注结果。
% 具体而言，门控网络对工人行为的映射，将专家模块的输出加权求和,然后通过一个全连接ReLU层和一个Softmax层，最终得到模型中工人$w_j$对item $x_i$的标注输出：
% \begin{equation}
%     f_{\bm{\Theta}} (\langle x_i, w_j\rangle )=\bm{o}\left(proj_{\bm{U}(x_i)} (w_j)\right)，
% \end{equation}
% 其中$\bm{o}$是输出函数，$\bm{\Theta}$是MMLC模型中$\bm{U}$, $\bm{v}$, $\bm{o}$这三个函数的参数集。
% 模型MMLC求解的问题是一个包含 $|\mathcal{K}|$ 个类别的分类问题，所以网络的输出是一个$|\mathcal{K}|$维向量，每个元素表示该类别的预测概率。
%%%%%%%%%%%%%%%%%%%%%%%%%%%%%%%%%%%%%%%%%
In the output module, the worker's labels for the item are generated. The output module generates labels for each worker based on their chosen expert modules. It involves mapping worker behavior through the gate network, which includes weighting and summing the outputs of the expert modules. Subsequently, through a fully connected ReLU layer and a softmax layer, the model produces the label output of worker $w_j$ for item $x_i$ as follows:
\begin{equation}
    f_{w_j} (x_i|\bm{\Theta})=\bm{o}\left(proj_{\bm{U}(x_i)} (w_j)\right),
\end{equation}
where $\bm{o}(\cdot)$ denotes the output function, and $\bm{\Theta}$ is the parameter set of the functions $\bm{U}$, $\bm{v}$, and $\bm{o}$ within the MMLC model. The MMLC model deals with a classification problem with $|\mathcal{K}|$ categories. The network's output is a $|\mathcal{K}|$-dimensional vector, where each element represents the predicted probability of a category.

%%%%%%%%%%%%%%%%%%%%%%%%%%%%%%%%%%%%%%%%%
% 模型的损失函数我们采用交叉熵损失函数加正则项的形式。具体损失函数为：
% \begin{equation}
%     \mathcal{L}_{\bm{\Theta}} = -\frac{1}{|\mathcal{D}|}\sum_{w_j\in \mathcal{W}}\sum_{x_i\in \mathcal{X}_j}\sum_{k\in \mathcal{K}} y_{ij}^{k}\log \left(f_{\bm{\Theta}} (\langle x_i, w_j\rangle )\right)+ \lambda \|\bm{\Theta}\|_F,
% \end{equation}
% 其中，第一项是多分类交叉熵损失，第二项是模型参数集$\bm{\Theta}$的L2正则化，防止模型过拟合，$\lambda$是正则化系数，$\|.\|_{F}$表示Frobenius范数。
% 通过最小损失函数，我们可以得到最终的模型$\mathcal{M}^{*}:f_{\bm{\Theta}^*}(\cdot)$。它使用函数 $f_{\bm{\Theta}^*}(\langle x_i,w_j\rangle )$ 来生成或预测工人$w_j$对item $x_i$的标注，其中 $\bm{\Theta}^{*}$ 是通过优化得到的最佳参数。
%%%%%%%%%%%%%%%%%%%%%%%%%%%%%%%%%%%%%%%%%
The model's loss function combines a cross-entropy loss term with a regularization term. The loss function is formulated as follows:
\begin{equation}
    \mathcal{L}_{\bm{\Theta}} = -\frac{1}{|\mathcal{D}|}\sum_{w_j\in \mathcal{W}}\sum_{x_i\in \mathcal{X}_j}\sum_{k\in \mathcal{K}} y_{ij}^{k}\log \left(f_{w_j} (x_i|\bm{\Theta})\right)+ \lambda \|\bm{\Theta}\|_F,
\end{equation}
The first term denotes the multi-class cross-entropy loss, while the second term represents the regularization of the model's parameter set $\bm{\Theta}$ to prevent overfitting. In the equation, $\lambda$ is the regularization coefficient, and $\|\cdot\|_{F}$ denotes the Frobenius norm. By minimizing the loss function, we can obtain the final model $\mathcal{M}^{*}:f(\cdot|\bm{\Theta}^*)$. This model uses the function $f_{w_j}(x_i|\bm{\Theta}^*)$ to predict the labels of worker $w_j$ for item $x_i$, where $\bm{\Theta}^{*}$ represents the optimized parameters.

\noindent\textbf{MMLC with Oracle Worker Finding (MMLC-owf):}
%%%%%%%%%%%%%%%%%%%%%%%%%%%%%%%%%%%%%%%%%
% 从MMLC模型的介绍可以看出，其不能直接生成推理的真值，为了解决这个问题，我们这里采用从工人spectral space找寻oracle worker's projection vector来进行真值推理。具体而言，任何一个工人从理论上来讲都在工人spectral space中投影出一个向量，代表工人的特征，工人在spectral space形成散点分布，我们假设存在一个全知全能的oracle worker，那么这个工人理论上在spectral space有一个projection vector，且在MMLC模型的输出就是item的真值。由此，我们只要找到oracle worker在工人spectral space的projection vector，那么我们就可以将其输出作为真值推理的结果。如果我们把任意一个工人看作是oracle worker的一种随机错误表达，那么工人投影到spectral space的分布中心就是orcale worker的projection vector,即：
% \begin{equation}
%     \bm{v}_{oracle} = \bm{\tau}\left(\bm{v} (\mathcal{W})\right),
% \end{equation}
% 其中$\bm{\tau}(\cdot)$是找寻分布中心的函数，这里我们可以采用核密度，平均数、中位数等方法。
% 基于MMLC模型的Oracle Worker Finding方法（MMLC-owf）的关于item $x_i$真值推理结果为：
% \begin{equation}
%     f_{\bm{\Theta}^*} (\langle x_i, w_{oracle}\rangle ) = \bm{o}\left(\bm{U}(x_i) \bm{v}_{oracle}\right).
% \end{equation}
%%%%%%%%%%%%%%%%%%%%%%%%%%%%%%%%%%%%%%%%%
The MMLC model does not directly generate the ground truth for inference. To address this issue, this section proposes a method for inferring the ground truth by identifying the oracle worker's projection vector in the worker spectral space. 
Specifically, each worker is theoretically associated with a projection in the worker spectral space, representing their unique characteristics. Workers are distributed in the spectral space. We assume the existence of an omniscient oracle worker who possesses a projection vector in the spectral space and is capable of providing the ground truth in the MMLC model. Therefore, by identifying the projection vector of the oracle worker in the worker spectral space, we can consider its output as the inferred truth.
If we treat any worker as a random expression of the oracle worker's error, then the center of the worker's distribution projected onto the spectral space can be regarded as the projection vector of the oracle worker, that is,
\begin{equation}
    \bm{v}_{oracle} = \bm{\tau}\left(\bm{v} (\mathcal{W})\right),
\end{equation}
where the function $\bm{\tau}(\cdot)$ is used to determine the distribution center, which can be found using methods such as kernel density estimation, mean, median, etc. According to the MMLC model, the outcome of the Oracle Worker Finding method (MMLC-owf) for inferring the ground truth of item $x_i$ can be expressed as follows:
\begin{equation}
    f_{w_{oracle}} (x_i|\bm{\Theta}^*) = \bm{o}\left(\bm{U}(x_i) \bm{v}_{oracle}\right).
\end{equation}

\noindent\textbf{MMLC with Data Filling (MMLC-df):}
In addition to the MMLC-owf method, we propose a method using data filling under the MMLC model called MMLC-df, which utilizes the sparsity of crowdsourced data. A new crowdsourced dataset $\mathcal{D}'$ is constructed through data filling as follows:
\begin{equation}
    \mathcal{D}' = \mathcal{D}\cup \left\{\langle x_i,w_j,\hat{y}_{ij}\rangle  \Big| \hat{y}_{ij} = f_{w_j}(x_i|\bm{\Theta}^*),I_{ij}=0\right\}.
\end{equation}
Subsequently, any learning-from-crowds method applied to this new crowdsourced dataset can infer higher-quality ground truth compared to that obtained from the original data.

\begin{figure*}[t]
    \centering
    \includegraphics[width=0.85\textwidth]{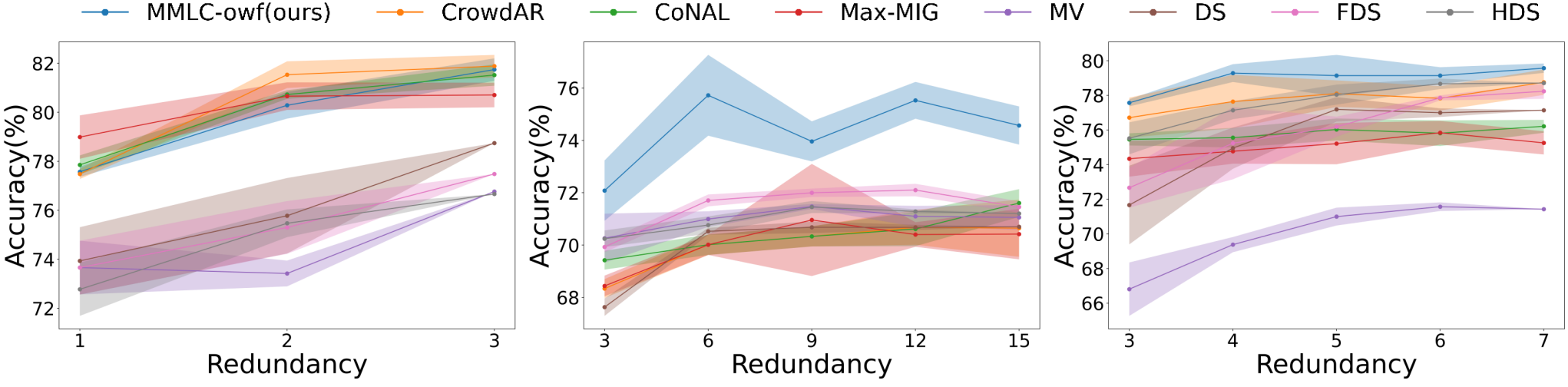}
    \caption{Accuracy Under Various Redundancies. (Left:\emph{LableMe}, Center:\emph{Text}, Right: \emph{Music})}
    \label{fig:redun}
\end{figure*}

\section{Experiments}\label{sec:experiments}

We verify the effectiveness of our method through experiments\footnote{Our code is available at https://github.com/Crowds24/MMLC.}.
We compare our learning-from-crowds method MMLC-owf with the following baselines: MV\cite{sheng2008get} directly uses majority voting to determine the ground truth; DS\cite{dawid1979maximum} employs a confusion matrix to characterize the labeling behavior of workers and uses the EM algorithm to infer the ground truth; HDS\cite{karger2011iterative} simplifies the DS method by assuming that each worker has the same probability of being correct under different truth values and equal probabilities for incorrect options; FDS\cite{sinha2018fast} is a simple and efficient algorithm based on DS, designed to achieve faster convergence while maintaining the accuracy of truth inference; Max-MIG\cite{cao2019max} utilizes the EM algorithm to integrate label aggregation and classifier training; CoNAL\cite{chu2021learning} distinguishes between common noise and individual noise by predicting a joint worker confusion matrix using classifiers; CrowdAR\cite{cao2023learning} estimates worker capability features through classifier prediction and models the reliability of joint worker labels.

We compare MMLC-df with the following baselines: G\_MV \cite{sheng2011simple} utilizes truth inference results from the MV algorithm to evaluate worker ability and assign new labels accordingly; G\_IRT \cite{baker2017basics} utilizes joint maximum likelihood estimation to estimate parameters of the IRT model, such as worker abilities and item difficulties, and generates new labels based on these parameters; TDG4Crowd \cite{fang2023tdg4crowd} learns the feature distributions of workers and items separately using worker models and item models. An inference component is used to learn the label distribution and generate new labels.

We identified three representative datasets that exemplify different types of crowdsourcing scenarios and data characteristics.
\emph{LableMe}~\cite{rodrigues2018deep,russell2008labelme}: This dataset consists of 1000 images categorized into 8 classes, with a total of 2547 labels provided by 59 workers. Each image is represented by 8192-dimensional features extracted using a pre-trained VGG-16 model. \emph{Text}~\cite{dumitrache2018crowdsourcing}: This dataset comprises 1594 sentences extracted from the CrowdTruth corpus, categorized into 13 groups. The dataset includes 14,228 labels provided by 154 workers. Each sentence is represented by 768-dimensional features extracted using a pre-trained BERT model.
\emph{Music}~\cite{rodrigues2014gaussian}: This dataset consists of 700 music compositions, each with a duration of 30 seconds, and categorized into 10 groups. It includes 2,945 labels provided by 44 workers. Each music composition is represented by 124-dimensional features extracted using the Marsyas \cite{rodrigues2013learning} music retrieval tool.

To accommodate the feature scales of the three experimental datasets, our model's architecture varies accordingly. For the \emph{LableMe} dataset, our model employs 16 expert modules, each comprising 3 fully connected ReLU layers, with a final layer output dimension of 32. For the \emph{Text} and \emph{Music} datasets, we utilize 10 expert modules. Each module consists of 3 and 2 fully connected ReLU layers, with output dimensions of 32 and 16, respectively. We adopt the settings from the learning-from-crowds methods Max-MIG, CoNAL, and CrowdAR, we adopt the settings from their source code for the \emph{LableMe} and \emph{Music} datasets. Since there is no source code available for the \emph{Text} dataset, we adopt the settings used in the \emph{LableMe} dataset. The hyperparameters mainly follow the expert configuration from Google's MMoE model\cite{ma2018modeling} and the classifier setup of CrowdAR. Regarding the TDG4Crowd data filling algorithm, we utilize the settings from its source code. The remaining methods do not use deep network structures and rely on default settings.

\begin{table}[tp]
\centering
\begin{tabular}{cccc}
\toprule
Method & \emph{LableMe} & \emph{Text} & \emph{Music} \\
\midrule
MV & 76.76 & 71.33 & 71.42 \\
DS & 79.73 & 70.70 & 77.14 \\
FDS & 77.78 & 71.45 & 77.57 \\
HDS & 76.66 & 71.20 & 78.01 \\
Max-MIG & 80.02$\pm$0.68 & 70.31$\pm$0.23 & 74.22$\pm$0.57 \\
CoNAL & 81.46$\pm$0.49 & 72.75$\pm$0.49 & 76.02$\pm$0.36 \\
CrowdAR & \textbf{82.14$\pm$0.36} & 70.48$\pm$0.42 & 78.54$\pm$0.59 \\
MMLC-owf & 81.74$\pm$0.47 & \textbf{74.31$\pm$0.39} & \textbf{79.14$\pm$0.21} \\
\bottomrule
\end{tabular}
\captionof{table}{Accuracy of Learning-from-Crowds Methods on Three Crowdsourced Datasets.}
\label{tab:accuracy} 
\end{table}

\begin{figure}
    \centering
    \includegraphics[width=0.48\textwidth]{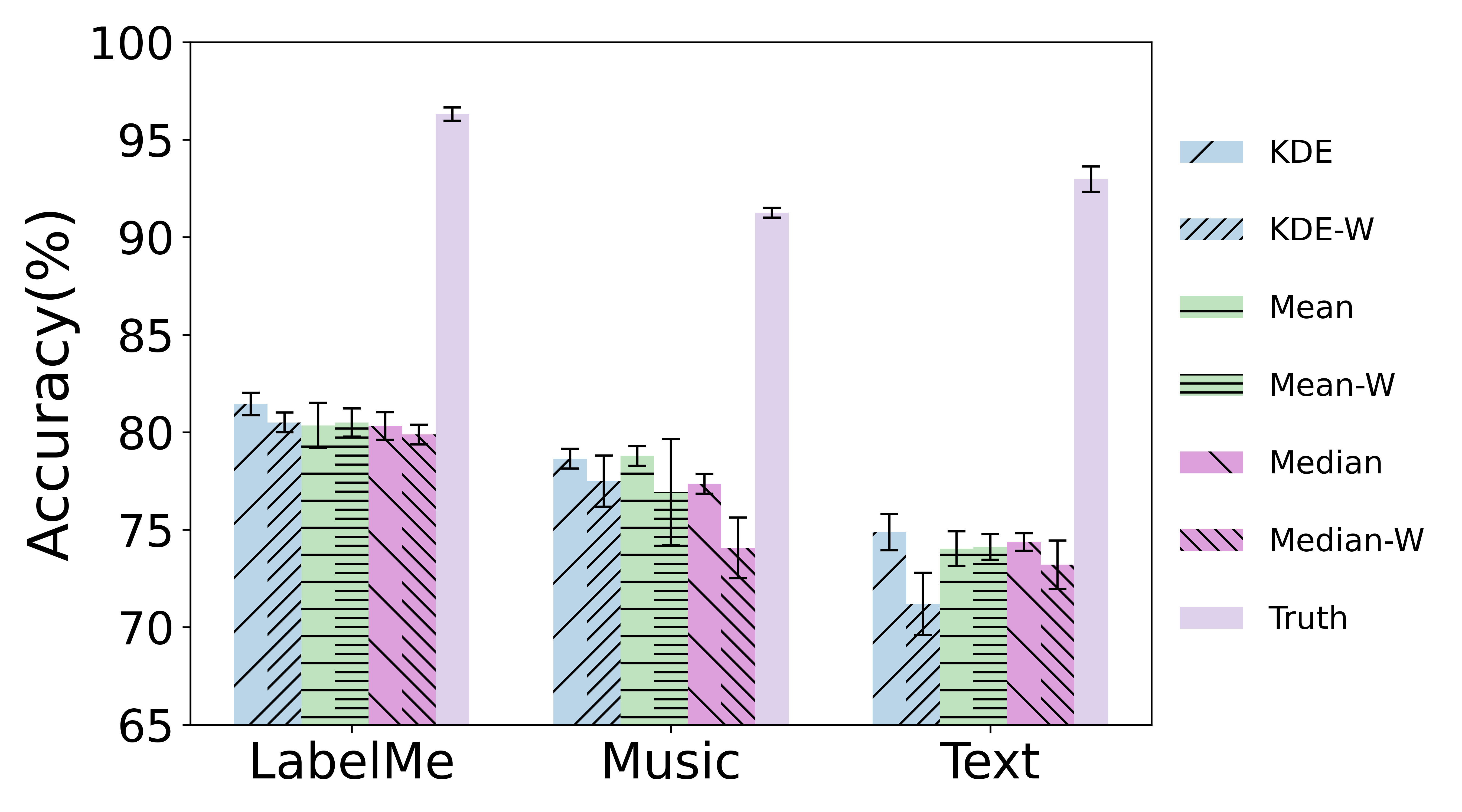}
    \caption{Accuracy of MMLC-owf with Various Clustering Methods on Three Crowdsourced Datasets.}
    \label{fig:oral}
\end{figure}

\begin{figure*}[ht]
    \centering
    \subfigure{
    \includegraphics[width=0.28\textwidth, height=0.18\textwidth]{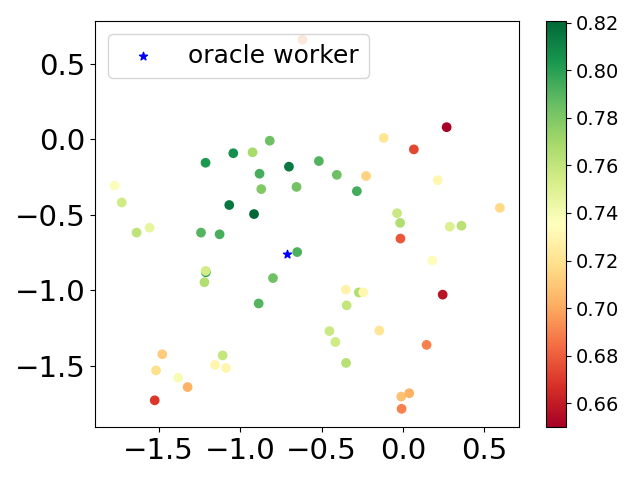}
    }
    \subfigure{
    \includegraphics[width=0.28\textwidth, height=0.18\textwidth]{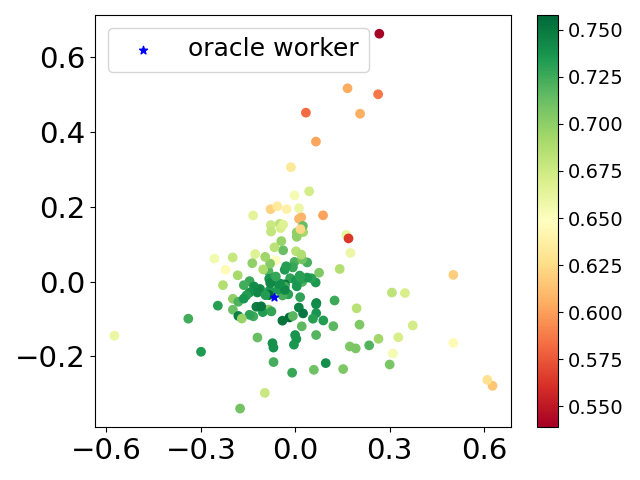}
    }
    \subfigure{
    \includegraphics[width=0.28\textwidth, height=0.18\textwidth]{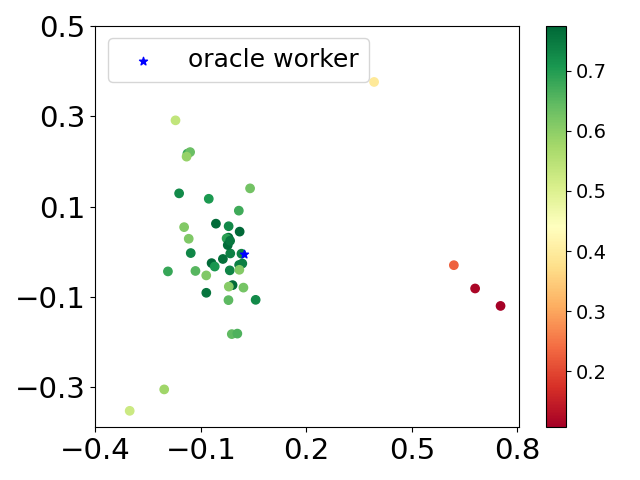}
    }
    \caption{Worker Scatter Plot in Worker Spectral Space.(Left:\emph{LableMe}, Center:\emph{Text}, Right: \emph{Music})}
    \label{fig:scatter}
\end{figure*}

\subsection{Evaluation of Oracle Worker Finding(MMLC-owf)}\label{sec:exp of MLC-ow}

\subsubsection{Main Result:}Our method, MMLC-owf, was evaluated alongside seven other methods through five rounds experiments. The average accuracy results are shown in Tab.~\ref{tab:accuracy}. In our method, we utilized kernel density estimation (KDE) to compute the projection vector of the oracle worker in the worker spectral space.
Our method, MMLC-owf, achieved the highest accuracy in the \emph{Text} and \emph{Music} datasets. In the \emph{LableMe} dataset, it ranked second, with only a 0.4\% difference from the top-performing CrowdAR method. Deep learning-based methods typically produce better results when analyzing datasets with high-dimensional item features like \emph{LableMe}. In datasets with fewer features, the advantage of deep learning methods was not significant.

\subsubsection{Impact of Redundancy:} We examine how varying levels of redundancy affect the accuracy of our method. Due to the varying redundancy of data items, a maximum redundancy parameter $R$ is set. We randomly keep $R$ labels for items with more than $R$ labels and discard the rest. This process generates a dataset with a maximum redundancy $R$. By conducting five repeated experiments and averaging the accuracy and standard deviation, the results are shown in Fig.~\ref{fig:redun}. As the average number of worker responses increases, all methods show an upward trend in their results. The analysis of various redundancy levels across the datasets indicates that higher redundancy levels are more advantageous for our method. Our method can effectively utilize worker behavior descriptions on datasets with higher redundancy but may face underfitting on datasets with lower redundancy.

\subsubsection{Clustering Methods in Oracle Worker Finding:}
Our method, MMLC-owf, utilizes a clustering method to determine the center of the distribution of the projection vector of workers in the worker spectral space as the projection vector of the oracle worker. Here, we examine how different clustering methods affect learning-from-crowds outcomes. We compare three clustering methods: kernel density estimation (KDE), Mean, and Median, as well as their worker-weighted variants: KDE-W, Mean-W, and Median-W. Worker weights are calculated based on the proportion of items answered by each worker relative to the total number of items, considering data imbalance.
In addition, The parameters of the expert modules and output modules are fixed and we optimize the projection vector in the worker spectral space using the ground truth of the items as the best performance of our model for clustering. This result is referred to as ``Truth.'' By conducting five repeated experiments and averaging the results, as shown in Fig.~\ref{fig:oral}. 
For example, in the \emph{LableMe} dataset, the oracle worker's projection vector is derived using KDE, KDE-W, Median, Median-W, Mean, and Mean-W. The MMLC-owf uses the oracle worker to generate ground truth. The quality of the ground truth obtained by the following five methods in each dataset is very similar, but the oracle worker generated using KDE produces the highest quality ground truth in each dataset. The ``Truth" method, which represents the theoretical upper limit with clustering methods, achieved accuracy rates of 96.32\%, 91.25\%, and 92.97\% on the three datasets respectively.The quality of the ground truth generated by oracle workers using six clustering methods still slightly deviates from theoretical upper limits. This implies that the MMLC-owf method can provide high-quality ground truth by optimally projecting the worker spectral space, approaching the theoretical upper limit. The model has strong expressive ability, with a small gap between the projected spectral space and the ground truth. There is potential to enhance MMLC-owf by choosing a more effective clustering method.

\subsubsection{Worker Distribution in Worker Spectral Space:} 
%We assume that any worker represents an oracle worker with random errors in their expression. Therefore, the center of the distribution of workers projected onto the spectral space corresponds to the projection vector of the oracle worker.
%We calculated the accuracy of each worker on the dataset, where higher accuracy is represented by a tendency towards green in the plot, while lower accuracy tends towards red.
%The projection obtained by the KDE method for the oracle worker is also presented in the plot, depicted by blue asterisks.
%From the distribution of worker projections, although the shapes of the distributions differ across datasets, there is a noticeable trend where workers with higher accuracy tend to cluster closer to the projection of the oracle worker. This observation demonstrates a clear tendency towards aggregation and provides some degree of confirmation for the validity of our assumption.
We assume that each worker is an oracle worker with random errors in their expression. The center of the distribution of workers projected onto the spectral space corresponds to the projection vector of the oracle worker.
To validate this assumption, we employed the IOSMAP dimensionality reduction method to reduce the worker projection vectors obtained from the MMLC model to 2D, resulting in the scatter plot shown in Fig.~\ref{fig:scatter}.
We calculated the accuracy of each worker on the dataset, where the closer the point's color on the graph is to green, the worker's accuracy is higher. The closer the point's color is to red, the lower the worker's accuracy.
The plot also shows the projection obtained by the KDE method for the oracle worker, represented by blue asterisks.
From the distribution of worker projections, although the shapes of the distributions differ across datasets, there is a noticeable trend where workers with higher accuracy tend to cluster closer to the projection of the oracle worker. This observation demonstrates a clear tendency towards aggregation and provides some degree of confirmation for the validity of our assumption.

\begin{table*}[ht]
\centering
\begin{tabular}{ccccccc}
    \toprule
     Data & Method & Original & G\_MV & G\_IRT & TDG4Crowd & MMLC-df\\
    \midrule
     & MV & 76.76 & \underline{-0.03$\pm$0.41} & \underline{-2.87$\pm$0.58} & +1.88$\pm$0.41 & \textbf{+3.89$\pm$0.17}\\
     & DS & 79.73 & \underline{-2.87$\pm$0.77} & +0.06$\pm$0.26 & \underline{-1.05$\pm$0.37} & \textbf{+1.72$\pm$0.20}\\
     & FDS & 77.78 & +0.07$\pm$0.92 & +0.20$\pm$0.62 & +0.93$\pm$0.23 & \textbf{+3.38$\pm$0.21}\\
    \emph{LableMe} & HDS & 76.66 & \underline{-0.23 $\pm$0.44} & +0.18$\pm$0.33 & +2.02$\pm$0.32 & \textbf{+2.48$\pm$0.36}\\
     & Max-MIG & 80.02$\pm$0.68 & +1.90$\pm$0.79 & +1.76$\pm$0.14 & +2.23$\pm$0.21 & \textbf{+4.81$\pm$0.21}\\
     & CoNAL & 81.46$\pm$0.49 & \underline{-1.92$\pm$0.41} & \underline{-0.78$\pm$0.68} & +0.02$\pm$0.72 & \textbf{+1.73$\pm$0.57}\\
     & CrowdAR & 82.14$\pm$0.36 & \textbf{+3.21$\pm$0.28} & +2.52$\pm$0.28 & \underline{-0.69$\pm$0.56} & +2.49$\pm$0.41\\
     & MMLC-owf & 81.74$\pm$0.47 & \underline{-2.91$\pm$0.46} & +0.05$\pm$0.61 & \underline{-1.41$\pm$0.54} & \textbf{+1.65$\pm$0.66}\\
    \midrule
     & MV & 71.33 & +0.53$\pm$0.31 & \underline{-0.26$\pm$0.44} & \underline{-0.01$\pm$0.08} & \textbf{+3.35$\pm$0.31}\\
     & DS & 70.72 & +1.80$\pm$0.53 & +0.01$\pm$0.51 & +0.38$\pm$0.21 & \textbf{+4.02$\pm$0.26}\\
     & FDS & 71.45 & +0.88$\pm$0.62 & \underline{-0.63$\pm$0.48} & \underline{-0.16$\pm$0.04} & \textbf{+3.26$\pm$0.35}\\
    \emph{Text} & HDS & 71.21 & +0.24$\pm$0.17 & \underline{-0.39$\pm$0.34} & \underline{-1.30$\pm$0.48} & \textbf{+2.18$\pm$0.19}\\
     & Max-MIG & 70.31$\pm$0.23 & \underline{-1.24$\pm$0.70} & \underline{-1.87$\pm$0.17} & +0.33$\pm$0.64 & \textbf{+3.88$\pm$0.51}\\
     & CoNAL & 72.75$\pm$0.49 & \underline{-1.33$\pm$0.22} & \underline{-2.27$\pm$0.43} & +0.62$\pm$0.32 & \textbf{+2.16$\pm$0.61}\\
     & CrowdAR & 70.48$\pm$0.42 & +0.37$\pm$0.62 & \underline{-0.33$\pm$0.21} & +1.96$\pm$0.12 & \textbf{+3.63$\pm$0.39}\\
     & MMLC-owf & 74.31$\pm$0.39 & \underline{-2.04$\pm$0.41} & \underline{-0.15$\pm$0.37} & +0.39$\pm$0.26 & \textbf{+1.46$\pm$0.52}\\
     \midrule
     & MV & 71.42 & \underline{-0.95$\pm$0.35} & +6.72$\pm$0.23 & +5.73$\pm$0.28 & \textbf{+7.58$\pm$0.45}\\
     & DS & 77.14 & \underline{-5.95$\pm$0.31} & +0.27$\pm$0.43 & +1.57$\pm$0.13 & \textbf{+2.41$\pm$0.41}\\
     & FDS & 77.57 & \underline{-6.47$\pm$0.64} & +0.41$\pm$0.57 & +0.81$\pm$0.08 & \textbf{+1.57$\pm$0.22}\\
    \emph{Music} & HDS & 78.01 & \underline{-7.25$\pm$0.87} & +0.52$\pm$0.43 & \textbf{+0.70$\pm$0.25} & +0.12$\pm$0.54\\
     & Max-MIG & 74.22$\pm$0.57 & +1.42$\pm$0.50 &+0.32$\pm$0.98 & \textbf{+5.54$\pm$0.72} & +4.77$\pm$0.41\\
     & CoNAL & 76.02$\pm$0.36 & \textbf{+7.97$\pm$0.54} & +4.65$\pm$0.54 & +5.55$\pm$0.51 & +6.37$\pm$0.66\\
     & CrowdAR & 78.54$\pm$0.59 & +1.46$\pm$0.42 & +2.31$\pm$0.20 & +1.97$\pm$0.11 & \textbf{+2.47$\pm$0.28}\\
     & MMLC-owf & 79.14$\pm$0.21 & +0.24$\pm$0.37 & +0.75$\pm$ 0.32 & \textbf{+1.62$\pm$0.31} & +1.43$\pm$ 0.42 \\
    \bottomrule
\end{tabular}
\caption{The Change of Accuracy After Data Filling.}\label{tab:fill}
\end{table*}

\begin{figure*}[ht]
    \centering
    \includegraphics[width=0.8\textwidth]{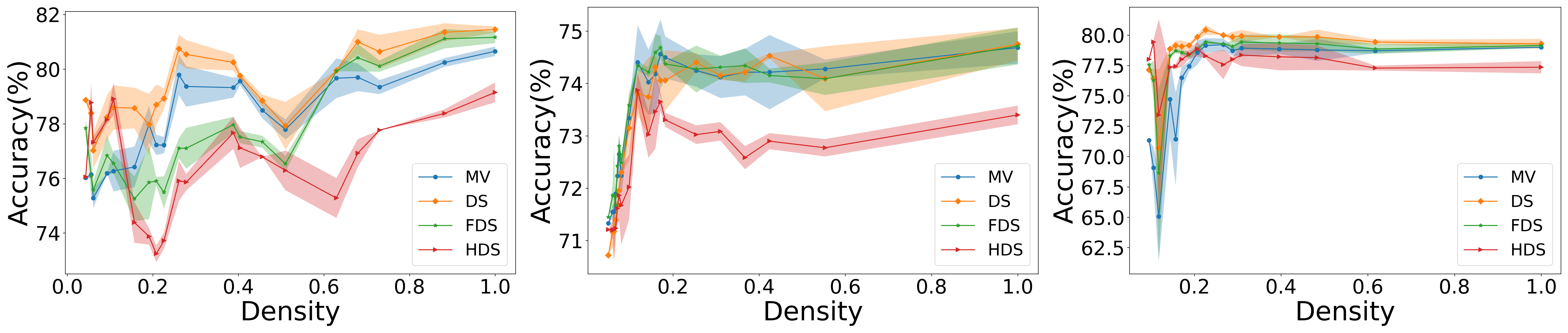}
    \caption{Accuracy with Various Data Filling's Density. (Left:\emph{LableMe}, Center:\emph{Text}, Right: \emph{Music})}
    \label{fig:select}
\end{figure*}

\subsection{Evaluation of Data Filling (MMLC-df)}

\subsubsection{Main Results:} 
%Our MMLC model constructs a worker behavior model that can directly generate crowdsourced data. Leveraging the sparsity of crowdsourced data, We propose a method based on data filling(MMLC-df). We compare our method with three filling methods, namely G\_MV, G\_IRT, and TDG4Crowd, and apply the filled data to the eight truth inference methods mentioned in this paper. By conducting five repeated experiments, the comparison results of accuracy are presented in Table \ref{tab:fill}.The table shows in three datasets, using eight truth inference methods across 24 scenarios.It can be observed that our MMLC-df framework achieves enhanced performance compared to the original data in 100\% of the scenarios, with 79.2\% of the scenarios achieving optimal enhancement. On the other hand, G\_MV, G\_IRT, and TDG4Crowd achieve enhanced results in 50\%, 62.5\%, and 75\% of the scenarios, respectively. In terms of the enhancement magnitude, our method performs the best on the \emph{Text} dataset. While other methods may achieve better results in certain scenarios for other datasets, their performance is unstable, and there are cases where the results deteriorate. This indicates that our MMLC-df framework demonstrates good stability and consistent performance.
We compared MMLC-df with three filling methods: G\_MV, G\_IRT, and TDG4Crowd. We used the filled data with eight learning-from-crowds methods from the previous section to infer the ground truth. We used three real datasets and applied eight truth value inference methods to infer the ground truth, resulting in a total of 24 scenarios. We conducted five rounds of experiments, and the mean and variance of all ground truth accuracy are presented in Tab.~\ref{tab:fill}. 
It can be observed that our MMLC-df framework achieves enhanced performance compared to the original data in 100\% of the scenarios, with 79.2\% of the scenarios achieving best enhancement. On the other hand, G\_MV, G\_IRT, and TDG4Crowd achieve enhanced results in 50\%, 62.5\%, and 75\% of the scenarios, respectively. In terms of the enhancement magnitude, our method performs the best on the \emph{Text} dataset. While other methods may achieve better results in certain scenarios for other datasets, their performance is unstable, and there are cases where the results deteriorate. This indicates that our MMLC-df framework demonstrates good stability and consistent performance.

\subsubsection{Impact of Data Filling's Density:}
%Our MMLC-df method leverages the sparsity of crowdsourced data for data filling. To better describe this, we define the data density of non-empty crowdsourced data $d_\mathcal{D}\in (0,1]$:
Our MMLC-df framework leverages the sparsity of crowdsourced data for data filling. To clarify, we define the data density of non-empty crowdsourced data as $d_\mathcal{D}\in (0,1]$ and $d_{\mathcal{D}} = \frac{|\mathcal{D}|}{|\mathcal{W}|\times |\mathcal{X}| }$.
The data densities of the three original datasets \emph{LabelMe}, \emph{Text}, and \emph{Music} are 0.0431, 0.0579, and 0.0956, respectively. The original data seems sparse. We gradually fill the data until reaching a data density to 1 for the analysis of its impact of data density on the results. We set a threshold for the number of items to be filled, denoted as $n_{interval}$. For workers with items exceeding this threshold, we replace their labeled items with predicted values. By adjusting the threshold from large to small, we gradually fill the data until all workers have completed their items, achieving a data density of 1.
Due to the large amount of filled data, deep learning methods can be time consuming. 
%Experiments have shown that and trends among various methods are quite similar.
The accuracies obtained by various methods show a similar trend to the density transformation. Therefore, we conducted this experiment using only statistical machine learning methods. The experiment is repeated in five rounds, and the average accuracy and standard deviation are shown in Fig.~\ref{fig:select}.
The trends are generally consistent across all methods, but the variations differ significantly among different datasets. In \emph{Text} dataset, as density increases, the algorithm's accuracy stabilizes rapidly and then reaches a plateau. In the \emph{LableMe} dataset, accuracy fluctuates significantly as density increases. Higher density often improves accuracy. In \emph{Music} dataset, as density increases, accuracy initially fluctuates rapidly before stabilizing. The \emph{Text} data filling performs the best, likely due to the larger scale of this dataset compared to the other two, resulting in a more significant impact.

\section{Conclusion}
This paper introduces a novel crowd-learning paradigm called MLC.
Within this paradigm, we propose a feature-level worker behavior model called MMLC. Based on this model, we develop two learning-from-crowds methods: MMLC-owf, which uses oracle worker finding, and a framework MMLC-df based on data filling. Experimental results demonstrate the superior performance of MMLC-owf compared to other methods. Furthermore, we assess the theoretical upper performance limit of the MMLC-owf method, demonstrating its potential to enhance clustering method selection and validate its strong performance. The experiments also validate the effectiveness and stability of the MMLC-df framework in enhancing learning-from-crowds methods through data filling. Furthermore, we observed that our model exhibited better performance on datasets with a higher number of annotations per worker. 
%On real crowdsourcing platforms, workers continuously engage in annotation tasks, resulting in an increasing average number of annotations per worker. Consequently, our model holds significant practical value for real world applications.
%\red{In future work, we will conduct detailed experiments on varying numbers of expert modules and analyze the behavioral changes of experts during training. This will further explore their specific impact on model performance and optimize the model design.}
\section{Acknowledgments}

This research has been supported by the Natural Science Foundation of Zhejiang Province, China (Grant No. LQ22F020002, LZ22F020008, and LQ24F020003), the National Natural Science Foundation of China under grant 61976187. Besides, the authors want to thank the anonymous reviewers for the helpful comments and suggestions to improve this paper.
\bibliography{aaai25}

\end{document}